\documentclass[journal]{IEEEtran} 
\usepackage{amsmath,amssymb,amsfonts,bm, mathtools, dsfont}
\usepackage{color}
\usepackage{cite}
\usepackage{textcomp}
\usepackage{xcolor}
\usepackage{amsthm}
\usepackage{multirow}
\usepackage{tabularray}
\usepackage{tikz}
\usepackage{booktabs}
\usepackage{siunitx}
\usepackage{pgfplots}
\usepackage{pgfplotstable}

\makeatletter
\pgfplotsset{compat=1.18}
\gdef\pgfplots@numplotsofactualtype{0}
\makeatother

\makeatletter
\let\pgfplots@countplots@advance\relax
\makeatother

\usepackage{float}
\usepackage[caption=false]{subfig}
\usepackage{makecell}
\usepackage{tikz}
\usetikzlibrary{matrix, positioning, intersections}
\usetikzlibrary{pgfplots.fillbetween}

\usetikzlibrary{backgrounds}

\usepackage{helvet}
\usepackage{stmaryrd}

\newtheoremstyle{boldtheorem}
  {3pt}   
  {3pt}   
  {}      
  {}      
  {\bfseries}  
  {.}     
  {1em}   
  {}      

\usepackage{hyperref}
\usepackage{scalerel}
\usepackage{algorithm}   
\usepackage{algorithmic}

\makeatletter
\newcommand\notsotiny{\@setfontsize\notsotiny\@vipt\@viipt}
\makeatother

\newcommand{\pacstl}{pacSTL}

\usepackage{acronym}
\acrodef{stl}[STL]{signal temporal logic}
\acrodef{colregs}[COLREG]{Convention on the International Regulations for Preventing Collisions at Sea}
\acrodef{tcpa}[TCPA]{time to closest point of approach}
\acrodef{pac}[PAC]{Probably Approximately Correct}
\title{\LARGE \bf
Staying on Spec: Real-Time Monitoring under \\ Uncertainty with a Maritime Case Study
}

\author{Anonymous Authors}
\author{Elizabeth Dietrich$^{\dagger, 1}$, Hanna Krasowski$^{\dagger, 1}$, Emir Cem Gezer$^{2}$, \\ Roger Skjetne$^{2}$, Asgeir Johan Sørensen$^{2}$, and Murat Arcak$^{1}$ 
\thanks{$^{\dagger}$Equal conribution}
\thanks{$^{1}$Elizabeth Dietrich, Hanna Krasowski, Murat Arcak are with the University of California, Berkeley
        {\tt\small \{eadietri, krasowski, arcak\}@berkeley.edu}}%
\thanks{$^{2}$Emir Cem Gezer, Roger Skjetne, Asgeir Sørensen are with the Norwegian University of Science and Technology
        {\tt\small \{emir.cem.gezer, roger.skjetne, asgeir.sorensen\}@ntnu.no}}%
        \vspace*{-2mm}
}

\begin{document}

\maketitle
\thispagestyle{empty}
\pagestyle{empty}

\begin{abstract}
Robotic systems must operate under uncertainty while satisfying complex task and safety specifications. Monitoring such specifications under uncertainty remains challenging, as existing 
formulations typically require extensive data or explicit uncertainty distributions.
In this paper, we propose a real-time monitoring framework that reduces data requirements by leveraging data-driven reachable sets for specification evaluation. 
We instantiate the framework for maritime navigation, where complex specifications arise from traffic rules. We develop a data-efficient pipeline for constructing reachable sets and derive a monitoring formulation suitable for real-time deployment. Simulation and hardware experiments demonstrate robust monitoring under realistic disturbances, achieving improved risk detection compared to state-of-the-art metrics.
\end{abstract}



\section{Introduction}

Temporal logic provides a flexible framework to encode and monitor complex behavioral requirements in many robotics applications \cite{Li2017, SrinivasanCoogan2021-TL2CBF-TRO, choe2025seeingsayingsolvingllmtotl}.
In maritime navigation, temporal logic has been used to falsify autonomous surface vessel controllers \cite{torben2023automatic, mueller2025falsificationdrivenRL}, monitor underwater vehicles \cite{Fossdal2024}, and certify surface vessel specification compliance \cite{Krasowski2024.safeRLautonomousVessels}. However, these approaches either neglect uncertainty \cite{mueller2025falsificationdrivenRL, Fossdal2024} or depend on simplified models to overapproximate system behavior \cite{Krasowski2024.safeRLautonomousVessels}, limiting their applicability in real-world settings where disturbances are pervasive. 

Current approaches typically address uncertainty in trajectory predictions through model-based reachability methods with bounded uncertainty estimation, where uncertainty is characterized through online data collection \cite{Mahesh2025, akhormeh2025onlinedatadrivenreachabilityanalysis} or historical data \cite{alanwar2022, laura2026}.
Resulting reachable sets are generally assumed to be valid over-approximations, which is difficult to verify when only limited data is available due to real-world resource constraints. In contrast, when generating data is inexpensive, e.g., in simulation, various data-driven reachability methods exist to directly compute tight reachable sets with probabilistic guarantees \cite{Devonport2020, Paccagnan2025-CDC, pmlr-v242-dietrich24a}. However, this typically results in a sim-to-real gap when deployed on hardware. 

\begin{figure}
\centering
        \input{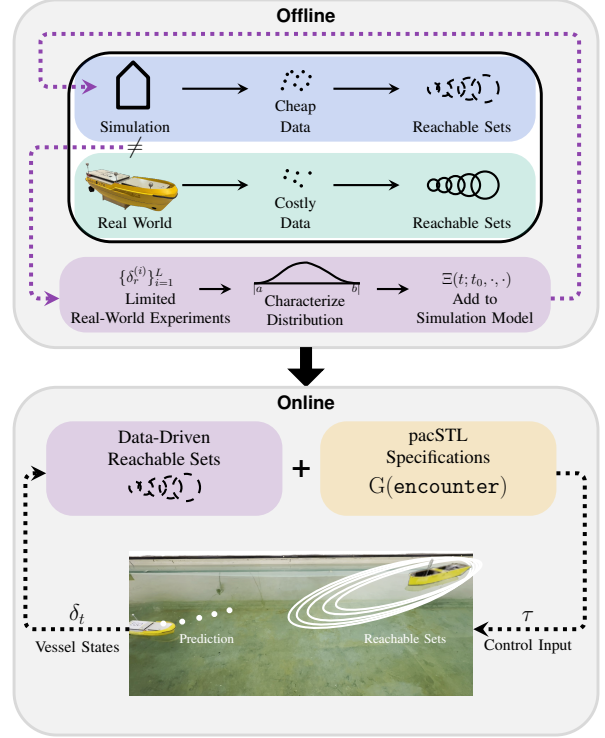}
    
    \caption{Offline: We use experimental data to characterize the disturbance distribution and incorporate it into the simulation model, bridging the sim-to-real gap (purple) and enabling inexpensive generation of real-world representative trajectories for reachability analysis. Online: The resulting reachable sets are combined with pacSTL specifications for real-time monitoring.}
    \vspace*{-2mm}
    \label{fig:new_headfig}
\end{figure}

In this paper, we present a real-time monitoring framework 
and demonstrate its use in maritime navigation 
(see Fig.~\ref{fig:new_headfig}). 
We build on the probabilistic signal temporal logic language, pacSTL \cite{krasowski2026pacstlpacboundedsignaltemporal}, and formulate realistic and computationally efficient monitoring specifications for maritime navigation, allowing us to account for uncertainty in real time. Additionally, we propose a real-world, data-driven reachability procedure that estimates distributional support from limited experimental data. We compare our framework against state-of-the-art baselines in hardware experiments with varying wave disturbances. 
Our main contributions include:

\begin{itemize}
    \item We introduce an efficient pacSTL formulation for maritime monitoring and decision making;  
    \item We propose an approach to data-driven reachability that integrates distribution modeling with high-fidelity simulations and leverages system invariances for deployment;
    \item We compare against \ac{stl} and a commonly used risk metric, demonstrating that pacSTL offers greater expressiveness and more robust monitoring, especially under disturbances, such as waves.
\end{itemize}

\section{Real-time Uncertainty-aware Monitoring}
Assessing temporal logic specification compliance under unbounded uncertainty is typically data-intensive \cite{sadigh2016safe,yoo2015control, TIGER2020325, Lin20203}. 
This is especially challenging for multi-agent relative states, as in maritime navigation.
To address this challenge, pacSTL~\cite{krasowski2026pacstlpacboundedsignaltemporal} was introduced as a probabilistic \ac{stl} language that evaluates specifications over reachable sets with \ac{pac} guarantees \cite{pacbound}. 

Given a \ac{pac}-bounded reachable tube, pacSTL computes lower $\underline{h}_t$ and upper $\overline{h}_t$ robustness bounds for each atomic proposition at every time step. Specifically, for a robustness function $h(\delta_t)$, where $\delta_t$ denotes the system trajectory at time $t$, pacSTL computes these bounds by solving optimization problems over the reachable set, $\mathcal{R}_t$: 
\begin{equation}
    \label{eq:lowbound}
        \underline{\mathrm{h}}_t = \underset{\delta_t
        }{\mathrm{minimize}} \;  h (\delta_t) \; \; \mathrm{subject \; to} \; \delta_t \in \mathcal{R}_t,
    \end{equation}
    \begin{equation}
    \label{eq:upbound}
        \overline{\mathrm{h}}_t = \underset{\delta_t}{\mathrm{maximize}} \;  h (\delta_t) \; \; \mathrm{subject \; to} \; \delta_t \in \mathcal{R}_t.
    \end{equation}
The resulting interval $[\underline{h}_t, \overline{h}_t]$ bounds the robustness of any state contained in $\mathcal{R}_t$ and inherits its \ac{pac} guarantee.
These atomic robustness intervals are used to recursively evaluate the \ac{stl} specification using interval semantics \cite{Baird2023}, obtaining a specification-level robustness interval and \ac{pac} bound on robustness interval satisfaction. We refer the reader to \cite{krasowski2026pacstlpacboundedsignaltemporal} for the details on syntax, semantics, and bound derivations.

In this work, we build on pacSTL to enable specification monitoring under uncertainty for maritime navigation.
Deploying pacSTL in the real world requires two key capabilities: (i) PAC-bounded reachable sets representative of real-world behavior, and (ii) a specification formulation that enables real-time evaluation. 
We address these challenges by developing a sim-to-real calibration procedure that aligns simulated trajectories with the real-world distribution using limited experimental data (Sec.~\ref{sec:data_driven_RA}), and by introducing efficiently evaluatable atomic propositions for maritime navigation (Sec.~\ref{sec:maritime_nav}). 
Together, these contributions facilitate real-time monitoring of complex temporal logic specifications under uncertainty using offline-computed reachable sets, without requiring a symbolic model of the system. 

\section{Data-Driven Reachability Analysis with Real-World Disturbances}\label{sec:data_driven_RA}

Data-driven reachability analysis enables reasoning about systems characterized by probability distributions over initial conditions, disturbances, or model parameters \cite{Devonport2020, pmlr-v242-dietrich24a, Paccagnan2025-CDC, devonport-2021-cdc}. Compared to model-based reachability \cite{Althoff2021, chen2018}, the data-driven paradigm is particularly useful when accurate models are unavailable, and system behavior can only be observed through simulation or experimentation.
Although data-driven methods can be applied directly to real-world data, collecting sufficient i.i.d. samples to obtain meaningful statistical guarantees is often time- and resource-intensive. 
Conversely, while simulations provide abundant data, they typically suffer from a sim-to-real gap.
To address these challenges, we propose a data-driven reachability framework (see Fig.~\ref{fig:new_headfig}) that combines data-driven disturbance modeling with sufficiently high-fidelity simulations. Experimental data is used to characterize a disturbance distribution, which is incorporated into a nominal simulation model to generate trajectories for reachable set computation. We formalize this approach below. 

A forward reachable set is defined as $\mathcal{R}_t = \{\Xi(t;t_0, x_0, d) : x_0 \in \mathcal{X}_0, d \in \mathcal{D}\}$ where $\mathcal{X}_0 \subseteq \mathbb{R}^{n_x}$ is the set of initial states, $\mathcal{D}$ is the set of disturbance signals $d: [t_0, t] \rightarrow \mathbb{R}^{n_d}$, and $\Xi(t;t_0, \cdot, \cdot) : 
\mathbb{R}^{n_x} \times \mathcal{D} \rightarrow \mathbb{R}^{n_x}$ is the state transition function. 
$\mathcal{R}_t $ contains all states that the system can transition to at time $t$ from $\mathcal{X}_0$ at $t_0$, subject to disturbances in $\mathcal{D}$.
Further, we denote a forward reachable tube as a collection of reachable sets $\mathcal{R} = \{\mathcal{R}_0, \dots, \mathcal{R}_T \}$. 
To estimate $\mathcal{R}_t$, we first specify the initial state set $\mathcal{X}_0$ and disturbance set $\mathcal{D}$, which may include disturbance terms, uncertain parameters, or other application-specific sources of variability. 
The set $\mathcal{X}_0$ is selected given the operating conditions of interest, while $\mathcal{D}$ is characterized from real-world experimental data $\{\delta_r^{(i)}\}^L_{i=1}$. 

Although the distribution of $\mathcal{D}$, $\mu_{\mathcal{D}}$, could ideally be estimated directly from representative, experimental data, obtaining sufficient i.i.d. data is often infeasible in practice.  
Given limited data and no prior distribution, the observed extrema of each disturbance component in $\{\delta_r^{(i)}\}^L_{i=1}$ can be used to parameterize a bounded support, $S_{\mathcal{D}}$, over which the disturbance distribution is modeled as uniform, i.e., $\mu_{\mathcal{D}} =\mathcal{U}(S_{\mathcal{D}})$. In the absence of additional distributional information, a uniform distribution is typically assumed \cite{jaynes1957}.

For estimating probabilistic reachable sets, let $\delta^{(i)} = \Xi(t; t_0, x_{0i}, d_i), i=1, \dots, N$ denote trajectories such that $x_{01}, \dots, x_{0N} \overset{i.i.d.}{\sim} \mu_{\mathcal{X}_0}$, $d_1, \dots, d_N \overset{i.i.d.}{\sim} \mu_{\mathcal{D}}$, where $\mathcal{X}_0$ is endowed with distribution $\mu_{\mathcal{X}_0}$, and $\mu_{\mathcal{D}}$ is as characterized above.
For a desired confidence parameter, $\beta$, we aim to find the minimum-volume $\mathcal{R}_t$ containing $\{\delta\}^N_{i=1}$, such that $ P^M (P_{\delta \in \mathcal{R}_{t}} \geq 1 - \epsilon_{\mathcal{R}_t} ) \geq 1 - \beta$ holds for $M$ i.i.d. samples, and the accuracy $\epsilon_{\mathcal{R}_t}$ is dependent on the verification approach. Note that due to the independence of samples $\delta^{(i)}$, $P^M$ denotes the product probability over $M$ samples, where $i=1,\dots,M$.

\section{Maritime Navigation}\label{sec:maritime_nav}
Maritime navigation is structured by traffic rules that describe proper maneuvers in case of a collision risk. More specifically, we focus on the traffic rules for power-driven vessels on the open sea described in the \ac{colregs} \cite{international1972convention}. Specifically, for power-driven vessels, there are three specified encounters (crossing, head-on, overtaking) and two collision-avoidance behaviors (give-way and stand-on). These encounters are always specified for two vessels, in our case, an autonomous ego vessel and another traffic participant, for which we can compute data-driven reachable sets. We denote a vessel trajectory as
$\delta_t\in \mathbb{R}^6$, consisting of surface position $\mathbf{p} = [p_x, p_y]^\top$, orientation $\psi$, velocity $\mathbf{v} = [v_x, v_y]^T$, and absolute velocity $vel = \| [v_x, v_y]\|_2$, where superscripts $E$ and $O$ denote ego and other vessel, respectively.

A head-on or crossing encounter is present when another vessel approaches the ego vessel from a specified sector and poses a risk of collision in the near future. A head-on encounter corresponds to an approach within the front sector (i.e., $\pm \SI{10}{\degree}$ from the orientation of the ego vessel), whereas a crossing encounter corresponds to an approach from the right. 
For the computational verifiability of these natural language rules, formalizations using temporal logic have been developed \cite{Krasowski2021.MarineTrafficRules, Krasowski2024.safeRLautonomousVessels, mueller2025falsificationdrivenRL, torben2023automatic,pedersen2025formalizing}.

\subsection{Atomic Propositions}\label{subsec:atomic_propositions}
We build on the formalizations in \cite{Krasowski2021.MarineTrafficRules, Krasowski2024.safeRLautonomousVessels, mueller2025falsificationdrivenRL} to specify robustness measures for atomic propositions that can be efficiently evaluated with \pacstl{}.
The maritime use case illustrates a realistic setting for evaluating different robustness measures as optimization objectives and demonstrates that efficient algorithms can be developed for nonlinear robustness measures.
Since the \ac{colregs} is specified between two vessels, the robustness measures of the atomic propositions are functions of signals from both the ego and other vessels. We adapt the robustness measures of the atomic propositions for maritime traffic rules from \cite[Sec. 5.2]{mueller2025falsificationdrivenRL}. 
Specifically, we build on the following atomic propositions:
\begin{itemize}
	\item $h_{\texttt{position\_halfplane}}$: linear function to detect relative positions
	\item $h_{\texttt{collision\_risk}}$: quadratic function to detect potential collisions
	\item $h_{\texttt{orientation\_halfplane}}$: nonlinear function to detect relative orientations
\end{itemize}

\paragraph{Linear Atomic Propositions}\label{subsec:lin_ap}

For linear atomic propositions, the robustness function is $h_\mathrm{lin} (\delta_t)= a^\top \, \delta_t - b$, where $a$ is a vector with the same dimensions as $\delta_t$, and $b$ is an offset. To calculate the maximum and minimum robustness, we solve \eqref{eq:lowbound} and \eqref{eq:upbound} with $h = h_\mathrm{lin}$. 
In maritime navigation, relative positions are specified by linear atomic propositions, which determine if the other vessel is in a position sector that may require evasive maneuvers. Specifically, we use the atomic proposition ${\mathtt{position\_halfplane}}$ presented in \cite{mueller2025falsificationdrivenRL}:
\begin{align}
    h_{\texttt{position\_halfplane}}&(\gamma^p, \sigma, \delta_t^E, \delta_t^O, v_\mathrm{max})= \\
    &\frac{\sigma}{v_{\mathrm{max}}}
    \begin{bmatrix}
    - \sin(\psi^E_t + \gamma^p) \\
    \cos(\psi^E_t + \gamma^p)
    \end{bmatrix}^\top  (\mathbf{p}^O_t - \mathbf{p}^E_t), \notag
\end{align}
where $\gamma^p$ defines one side of the position sector as a halfplane passing through the ego vessel position and is set according to the \ac{colregs}. The maximum velocity $v_\mathrm{max}$ scales the robustness magnitudes to a time domain, and $\sigma \in \{1, -1\}$ determines which side of the halfplane satisfies the atomic proposition.
Evaluating $h_{\texttt{position\_halfplane}}$ at the predicted ego state $\delta_t^E$ results in
\begin{align}
    a_\mathrm{pos} &= \frac{\sigma}{v_{\mathrm{max}}}\begin{bmatrix} - \sin(\psi^E_t + \gamma^p) \\
    \cos(\psi^E_t + \gamma^p)
    \end{bmatrix}, \\
    b_\mathrm{pos} &= a_\mathrm{pos}^\top \,\mathbf{p}^E_t
\end{align}
as parameters for $h_\mathrm{lin}$. Note that dependency on the ego vessel state results in a state-dependent robustness function.

\paragraph{Quadratic Atomic Propositions}\label{subsec:quad_ap}

For quadratic atomic propositions, the general robustness function is $h_\mathrm{quad} = \delta_t^\top Q \delta_t + a^\top \delta_t + b$.
In maritime navigation, we use a quadratic function to quantify collision risk over a given time horizon, conservatively relaxing the collision predicate from \cite{mueller2025falsificationdrivenRL}:
\begin{align}
    h_{\texttt{collision\_risk}}&(\delta_t^E, \delta_t^O, \kappa_t^{\Box}; a_{\max}, t_h) = \\
    & \frac{1}{a_{\max}}\Big(\frac{\| \kappa_t^{\Box} \|_2}{t_h} - \|\mathbf{v}^E_t - \mathbf{v}^O_t \|_2\Big), \notag
\end{align}
where $a_{\max}$ is the maximum acceleration of the ego vessel and $\kappa_t^{\Box} = \underset{\mathbf{p}^O_t}{\Box}(\mathbf{p}^E_t - \mathbf{p}^O_t)$ is the minimal or maximal relative position between the ego vessel and the other vessel, with $\Box$ denoting $\min$ or $\max$, respectively. As a result, the robustness function is quadratic in $\delta^O$. We define the following optimization problems to compute the robustness bounds:
\begin{align}
\underline{\mathrm{h}}_t =& \underset{\mathbf{v}^O_t \in \mathcal{R}}{\text{minimize}} \; h_{\texttt{collision\_risk}} (\delta_t^E, \delta_t^O, \kappa_t^{\max}), \\
\overline{\mathrm{h}}_t =& \underset{\mathbf{v}^O_t \in \mathcal{R}}{\text{maximize}} \; h_{\texttt{collision\_risk}} (\delta_t^E, \delta_t^O, \kappa_t^{\min}).
\label{eq:upbound_quad_pred}
\end{align}

\paragraph{Special Atomic Propositions}\label{subsec:nonlin_ap}

\begin{figure}
    \centering
    \definecolor{pacwaves}{rgb}{0.992, 0.71, 0.082}
\definecolor{pacnowaves}{RGB}{0, 63, 195}
\definecolor{stlcolor}{HTML}{16A085}  
\definecolor{tcpacolor}{HTML}{8E44AD}  

\begin{tikzpicture}[
    dot/.style={circle, fill=black, inner sep=0pt, minimum size=4pt},
    label font/.style={font=\normalsize}
]

\def\R{2}

\fill[stlcolor, opacity=0.8] (0,0) -- (135:\R) arc (135:75:\R) -- cycle;
\fill[pacwaves, opacity=0.8] (0,0) -- (10:\R) arc (10:-35:\R) -- cycle;
\fill[tcpacolor, opacity=0.8] (0,0) -- (215:\R) arc (215:240:\R) -- cycle;

\draw[pacnowaves, very thick] (180:3.3) -- (0:3.3);

\draw[thick] (0,0) circle (\R);

\node[dot, minimum size=4.5pt] at (0,0) {};


\draw[thick] (0,0) -- (135:\R) node[dot] {} node[above left=1pt, label font] {\textcolor{stlcolor}{$\mathrm{\underline{h}}$}};
\draw[thick, dashed] (0,0) -- (90:\R) node[dot] {} node[above=2pt, label font] {\textcolor{stlcolor}{$\mathrm{\overline{h}}$}};
\draw[thick] (0,0) -- (75:\R) node[dot] {} node[above=2pt, label font] {\textcolor{stlcolor}{$\mathrm{h}^\prime$}}; 

\draw[thick] (0,0) -- (10:\R) node[dot] {} node[above right=1pt, label font] {\textcolor{pacwaves}{$\mathrm{\overline{h}}$}};
\draw[thick] (0,0) -- (-35:\R) node[dot] {} node[below right=1pt, label font] {\textcolor{pacwaves}{$\mathrm{\underline{h}}$}};

\draw[thick] (0,0) -- (215:\R) node[dot] {} node[above left=1pt, yshift=-0.4cm, label font] {\textcolor{tcpacolor}{$\mathrm{\overline{h}}$}};
\draw[thick] (0,0) -- (240:\R) node[dot] {} node[below left=1pt, label font] {\textcolor{tcpacolor}{$\mathrm{\underline{h}}$}};

\draw[thin] (75:1.0) arc (75:0:1.0) node[midway, above, xshift=0.1cm] {$\alpha^\prime$};
\draw[thin] (90:1.5) arc (90:0:1.5) node[midway, above] {$\alpha$};

\begin{scope}[xshift=-0.7cm]
\draw[line width=1.8pt, draw=pacnowaves] (3.5, 0.7) -- (3.9, 0.7);
\draw[line width=1.8pt, draw=pacnowaves] (3.7, 0.5) -- (3.7, 0.9);

\draw[line width=2.5pt, draw=pacnowaves] (3.5, -0.4) -- (3.9, -0.4);
\end{scope}

\end{tikzpicture}
    \caption{Illustrative computation of three $\mathtt{orientation\_halfplane}$ robustness intervals. The colored sectors represent orientation intervals projected from reachable sets, and the blue line denotes the robustness satisfaction boundary. The yellow example illustrates an orientation interval that overlaps the satisfaction boundary. The purple example shows the case in which the lower and upper robustness bounds are reversed.
    In the green example, $\mathrm{\overline{h}}$ is determined by the point $\SI{90}{\degree}$ from the satisfaction boundary because $\alpha = \SI{90}{\degree} > \alpha^{\prime}$, where $\alpha$ is the relative orientation corresponding to $\underline{\psi}_t^O$. }
    \vspace*{-2mm}
    \label{fig:orientation_halfplane}
\end{figure}
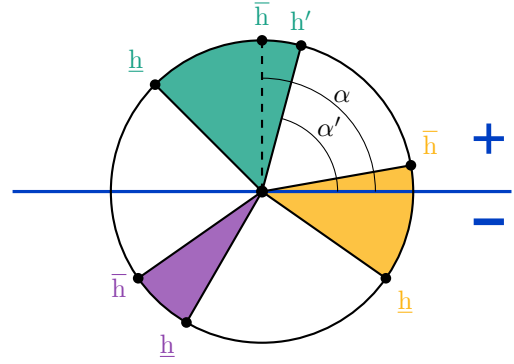

Solving \eqref{eq:lowbound} and \eqref{eq:upbound} is challenging for general nonlinear atomic propositions. 
However, efficient computation of lower and upper robustness bounds remains possible when there is only one signal relevant for the atomic proposition, or the nonlinear function can be tightly bound by linear functions. 
This setting is common in maritime navigation, where nonlinear atomic propositions with only one signal are often used to characterize the relative orientation between two vessels.
Specifically, for convex reachable sets, the relative orientation of the other vessel is bounded by a one-dimensional interval, enabling a case-wise computation of the lower and upper robustness measures:
\begin{align}
    &[\underline{\mathrm{h}}_t, \overline{\mathrm{h}}_t] = \\ &[\mathtt{orientation\_halfplane}](\gamma^\psi, \sigma, [\underline{\psi}_t^O , \overline{\psi}_t^O],\psi^E_t,  r_\mathrm{max}), \notag
\end{align}
where $[\underline{\psi}_t^O , \overline{\psi}_t^O]$ is the orientation interval at time step $t$, $\gamma^\psi$ is the relative orientation threshold, $\sigma \in \{1, -1\}$ determines if the threshold is a lower or upper bound, and $r_\mathrm{max}$ is the ego vessel's maximal angular velocity.
The orientation interval $[\underline{\psi}_t^O , \overline{\psi}_t^O]$ is calculated relative to the orientation threshold, defined by $\gamma^\psi$ and $\psi^E_t$ (see \cite[Fig. 3(b)]{mueller2025falsificationdrivenRL}). If any orientation ${\psi}_t^O \in [\underline{\psi}_t^O , \overline{\psi}_t^O]$ is more than $\pi/2$ from the satisfaction thresholds, we instead use the remaining angular distance to $\pi$.
Fig.~\ref{fig:orientation_halfplane} provides an illustration of this computation.

\subsection{pacSTL Specifications}\label{subsec:monitored_spec}
Based on these robustness functions for atomic propositions, \pacstl{} specifications are constructed according to I-\ac{stl} syntax \cite{Baird2023}. 
For maritime navigation, we present two encounter specifications that determine whether an evasion maneuver is required.  
Given the robustness functions from Sec.~\ref{subsec:atomic_propositions}, we denote the corresponding predicates using typewriter font. 

First, we define predicates that determine whether the other vessel will occupy the relative position or orientation sector corresponding to each $\mathtt{encounter}$. Let $\underline{\gamma^p}$ and $\overline{\gamma^p}$ denote the lower and upper orientations of the specified position sector relative to the ego orientation, with $\underline{\sigma}=-1$ and $\overline{\sigma}=1$.
The relative position and orientation specifications are defined using $\mathtt{position\_halfplane}$ and $\mathtt{orientation\_halfplane}$:
\begin{align}
        &\mathtt{pos\_encounter} \eqqcolon  \\
        &\mathtt{position\_halfplane}(\underline{\gamma^{p,e}}, \underline{\sigma}, \delta_t^E, \delta_t^O, v_\mathrm{max}) \land \notag \\
        &\mathtt{position\_halfplane}(\overline{\gamma^{p,e}}, \overline{\sigma}, \delta_t^E, \delta_t^O, v_\mathrm{max}), \notag
\end{align}
and
\begin{align}
        &\mathtt{ori\_encounter} \eqqcolon  \\
        &\mathtt{orientation\_halfplane}(\underline{\gamma^{\psi,e}},  \underline{\sigma}, [\underline{\psi}_t^O , \overline{\psi}_t^O],\psi^E_t,  r_\mathrm{max}) \land \notag \\
        &\mathtt{orientation\_halfplane}(\overline{\gamma^{\psi, e}}, \overline{\sigma}, [\underline{\psi}_t^O , \overline{\psi}_t^O],\psi^E_t,  r_\mathrm{max}). \notag
\end{align}
The superscript $e$ denotes the encounter type and is instantiated as $H$ and $C$ for head-on and crossing, respectively. The corresponding parameters are set according to the \ac{colregs} (see Table~\ref{tab:parameters} for values).
Now, we formalize each encounter specification as a conjunction of $\mathtt{pos\_encounter}$, $\mathtt{ori\_encounter}$, and $\mathtt{collision\_risk}$.
Specifically, for a head-on encounter:
\begin{align}
        &\mathtt{head\_on} \eqqcolon  \\
        &\mathtt{pos\_head\_on} \land \mathtt{ori\_head\_on} \land \mathtt{collision\_risk} \notag,
\end{align}
and for a crossing encounter:
\begin{align}
        &\mathtt{crossing} \eqqcolon  \\
        &\mathtt{pos\_crossing} \land \mathtt{ori\_crossing} \land \mathtt{collision\_risk}. \notag
\end{align}
The above predicates only capture instantaneous encounter conditions using Boolean operators. To detect if an evasive maneuver is necessary, we require these encounters to persist over a specified time interval $[t_\mathrm{start}, t_\mathrm{end}]$:
\begin{equation}\label{eq:persistenthead}
    \Phi_H \eqqcolon \mathrm{G}_{[t_\mathrm{start}, t_\mathrm{end}]} (\mathtt{head\_on}),
\end{equation}
\begin{equation}\label{eq:persistentcross}
    \Phi_C \eqqcolon \mathrm{G}_{[t_\mathrm{start}, t_\mathrm{end}]} (\mathtt{crossing}),
\end{equation}
where $\mathrm{G}$ is the temporal operator for always.

\section{Experimental Setup}
\label{sec:setup}

We evaluate the \pacstl{} formulation in simulation and on a physical vessel testbed, comparing against existing baselines.~\footnote{For reproducibility, we provide our implementation including parameters at: \url{https://anonymous.4open.science/r/pacSTLMaritime-4484/}}
This section describes the reachable set computation, navigation scenarios, experimental setup, and baseline implementations. Key parameters are summarized in Table~\ref{tab:parameters}.

\begin{table}[tb]
\caption{Evaluation Parameters}
\vspace{-1em}
\begin{center}
\footnotesize
\begin{tabular}{ll|ll}
\toprule
\textbf{\makecell[l]{Vessel\\Parameters}} & \textbf{\makecell[l]{\\Value}} & \textbf{\makecell[l]{Traffic Rule\\Parameters}} & \textbf{\makecell[l]{\\Value}} \\
\midrule
S/L Width& $\SI{0.3}{\meter} \, / \, \SI{0.4}{\meter}$ & $t_\mathrm{start}$ & $\SI{1.0}{\second}$\\
S/L Length & $\SI{1.0}{\meter} \, / \, \SI{2.6}{\meter}$ & $t_\mathrm{end}$ & $\SI{2.5}{\second}$\\
S/L Draft& $\SI{0.08}{\meter} \, / \,\SI{0.02}{\meter}$ & $\Delta t$ & $\SI{0.5}{\second}$\\
$v_\mathrm{max}$ & $\SI{0.4}{\meter\per\second}$ & $\underline{\gamma}^{p,H},\overline{\gamma}^{p,H}$ &  [\SI{10}{\degree}, \SI{-10}{\degree}] \\
$a_\mathrm{max}$ & $\SI{0.15}{\meter\per\second\squared}$ & $\underline{\gamma}^{p,C}, \overline{\gamma}^{p,C}$ &  [\SI{-10}{\degree},\SI{-112.5}{\degree}]\\
$r_\mathrm{max}$ & $\SI{0.8}{\radian\per\second}$ & $\underline{\gamma}^{\psi,H},\overline{\gamma}^{\psi,H}$ &  [\SI{170}{\degree},\SI{-170}{\degree}] \\
$v_\mathrm{des}$ & $\SI{0.1}{\meter \per \second}$& $\underline{\gamma}^{\psi,C},\overline{\gamma}^{\psi,C}$ &  [\SI{170}{\degree},\SI{10}{\degree}] \\
\bottomrule
\end{tabular}
\label{tab:parameters}
\vspace{-0.2cm}
\end{center}
\end{table}

\vspace{-2mm}
\subsection{Vessel Simulation Model}

We use a six-degree-of-freedom (6-DOF) vessel model that considers position and orientation in the inertial frame and velocities in the body frame \cite{fossen2011handbook}. We define $\eta = \begin{bmatrix}
    p_x, p_y, z, \phi, \theta, \psi
\end{bmatrix}^{\top}$ to be the vessel's position $(p_x, p_y)$, heave ($z$), roll $(\phi)$, pitch $(\theta)$, and yaw $(\psi)$ angles. Furthermore, $\nu = \begin{bmatrix}
    v_x, v_y, v_z, v_{\phi}, v_{\theta}, v_{\psi}
\end{bmatrix}^{\top}$ denotes the vessel's linear and angular velocities.
The dynamics for this system are: 
\begin{equation}
\label{eq:boat}
\begin{aligned}
    \dot{\eta} &= \mathbf{R}(\psi)\nu, \\
    \dot{\nu} &= \mathbf{M}^{-1}(\tau - \mathbf{C}(\nu)\nu - \mathbf{D}(\nu)\nu + b),
\end{aligned}
\end{equation}
where $\mathbf{R}(\psi)$ 
transforms velocities from the body-fixed frame to the world-fixed frame, as defined in \cite{fossen2011handbook}.
Further, $\mathbf{M} \in \mathbb{R}^{6\times6}$ is the inertia matrix, including the vessel’s mass and added mass terms, $\mathbf{C}(\nu) \in \mathbb{R}^{6\times6}$ represents Coriolis and centripetal forces, $\mathbf{D}(\nu) \in \mathbb{R}^{6\times6}$ is the hydrodynamic damping matrix accounting for drag forces, $\tau \in \mathbb{R}^6$ represents control forces and moments (e.g., thrusters), and $b \in \mathbb{R}^6$ represents slowly varying environmental loads or unmodeled dynamics.

\subsection{Reachable Set Computation}
To enable offline reachable set construction from simulation while maintaining consistency with real-world vessel behavior, we first characterize $\mu_{\mathcal{X}_0}$ and $\mu_{\mathcal{D}}$. 
In the maritime setting, disturbances are largely captured by
\begin{equation}
\label{eq:b}
    b = \mathbf{M}\dot{\nu} - \tau + \mathbf{C}(\nu)\nu + \mathbf{D}(\nu)\nu.
\end{equation}
Collecting sufficient real-world trajectories to directly construct statistically meaningful reachable sets is time- and resource-intensive. Therefore, we instead estimate the disturbance support $S_{\mathcal{D}}$ using data from approximately $100$ physical vessel trials conducted under representative operating conditions (e.g., still-water or wave conditions).
In each trial, the vessel is initialized at a deployment starting position and commanded to traverse the laboratory pool using randomly sampled control inputs $\tau$. Specifically, samples are drawn uniformly over feasible control actions based on vessel operating limits.
After each traversal, the vessel is reset to its initial positions and the process is repeated. Motion-capture measurements of vessel velocity $\nu$ and applied control inputs $\tau$ are used to compute the disturbance term $b$ from \eqref{eq:b}. Finally, the $b$-component of $S_{\mathcal{D}}$ is defined by the component-wise extrema of $b$ observed across all trials. Separate supports are estimated for still-water and wave conditions.

For reachable set computation, 
uncertainty in the control input $\tau$ is introduced as an additional disturbance.
Accordingly, the set of disturbance signals is 
\[
\mathcal{D} =
\left\{
\begin{aligned}
&\{b(t) = b, \hspace*{1mm} \forall t \in \{t_0, t_2, t_4, ... , t_T\}\} \; \cup \\
&\{\tau(t) = \tau, \hspace{1mm} \forall t \in \{t_0, t_1, t_2, ... , t_T\}  \}
\end{aligned}
\right\},
\]
where $b$ and $\tau$ are piecewise constant, and $\mu_{\mathcal{D}}$ is uniform over the identified support.
The different sampling frequencies of $b$ and $\tau$ reflect $b$'s dependency on the velocity derivative $\dot{\nu}$, making a factor-of-two scaling the smallest meaningful discretization interval.
In addition to $\mu_{\mathcal{D}}$, $\mu_{\mathcal{X}_0}$ is characterized to capture the vessel's operating conditions. The dynamics in \eqref{eq:boat} are invariant to global translations and equivariant under rotations, as they depend on relative orientation, velocity, inputs, and disturbances, rather than absolute position. We exploit this symmetry to reduce the number of reachable sets computed offline.
The initial-state set for surge velocity is partitioned into four intervals, $\{\mathcal{U}_i\}_{i=1}^{4}$, spanning the experimental operating speeds. 
Reachable sets are computed separately for each interval, with $\mu_{\mathcal{X}_0}$ taken as a uniform distribution over the corresponding velocity range. 

Using $\mu_{\mathcal{X}_0}$ and $\mu_{\mathcal{D}}$, we simulate trajectories of \eqref{eq:boat} and estimate reachable tubes $\mathcal{R}$ over $\delta_t\in \mathbb{R}^6$, consisting of the specification-relevant states $\mathbf{p}, \psi$, $\mathbf{v}$. 
While there are multiple approaches to obtain \ac{pac}-bounded reachable sets \cite{Devonport2020, pmlr-v242-dietrich24a, Paccagnan2025-CDC,devonport2023}, we employ sampling-based optimization to obtain ellipsoids \cite[Eq. (8)]{Devonport2020} and verify them using the holdout method \cite{dietrich2025datadrivenreachabilityscenariooptimization}. Ellipsoids provide compact set representations while remaining computationally tractable through efficient optimization formulations. 
For each of the four forward-velocity intervals, one reachable tube is constructed with prediction horizon $T = 2.5 \mathrm{s}$.
We use $N=1500$ i.i.d. samples for reachable set construction and $M=1500$ i.i.d. samples with $\beta=10^{-9}$ to compute probabilistic guarantees. To account for vessel geometry, the $x, y$ center coordinates of the reduced state are transformed into four corner points representing the vessel. 
We assume no distribution shift between disturbance characterization and deployment. 
The resulting reachable tubes $\mathcal{R}$ have accuracies $\epsilon \in [0.026, 0.045]$ in still-water and $\epsilon \in [0.029, 0.044]$ in waves, while the reachable sets $\mathcal{R}_t$ have $\epsilon \in [0.019, 0.039]$ and $\epsilon \in [0.018, 0.042]$, respectively.

\vspace{-3mm}
\subsection{Maritime navigation setting}

\begin{figure}[t]
   \centering
   \resizebox{.95\linewidth}{!}{
    \input{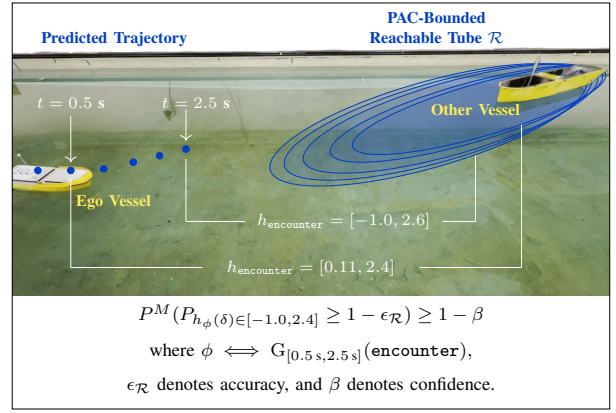}
    }
    \caption{Example evaluation \pacstl{} specification $\phi$, which is based on the example atomic proposition $\mathtt{encounter}$. 
    The resulting robustness interval of $\phi$, here $[-1.0, 2.4]$, is endowed with a probabilistic guarantee on the robustness value of an unseen trajectory being contained within the interval. The robustness intervals are used to monitor critical encounters and, consequently, trigger an evasive maneuver reliably for still-water and wave conditions.  }
    \label{fig:headfig}
    \vspace{-5mm}
\end{figure}

We demonstrate \pacstl{} monitoring for safe maneuvering on two scenarios, i.e., head-on and crossing, using a small vessel (S) and a large vessel (L). 
Each scenario consists of initial states for the ego $\delta_0^E$ and other vessel $\delta_0^O$, as well as a goal state for the ego vessel $\delta_\mathrm{goal}^E$. 
The parameters for specifications \eqref{eq:persistenthead} and \eqref{eq:persistentcross} are based on the \ac{colregs}. Note that the time parameters are adjusted to achieve realistic maneuvering in a confined lab space. 

During an experiment (see Fig.~\ref{fig:headfig}), the ego vessel monitors for possible encounters based on \pacstl{} specifications \eqref{eq:persistenthead} and \eqref{eq:persistentcross}, while using a line-of-sight (LOS) controller \cite{lekkas2013line} to steer towards 
the specified goal with a desired velocity $v_\mathrm{des} = \SI{0.1}{\meter \second^{-1}}$.
The ego vessel observes the current state of both vessels, predicts its own future states assuming constant speed and orientation, and selects reachable tube predictions given the other vessel's current state. 
The runtime for evaluating \pacstl{} specifications on hardware is on average 1.4Hz. 

Once the upper bound on robustness $\mathrm{\overline{h}}$ becomes positive, the ego vessel evades by altering the desired path.
Specifically, two additional waypoints are generated. The first waypoint is at the angle $\psi_\mathrm{turn} = \SI{0.8}{\radian}$ to the right of the ego vessel at a distance $d_\mathrm{turn} = v_\mathrm{des} t_\mathrm{turn}$, where $t_\mathrm{turn} = \SI{30}{\second}$ approximately specifies the time spent in the turning phase. The second waypoint is at the distance $d_\mathrm{parallel} =  v_\mathrm{des} t_\mathrm{parallel}$ from the first waypoint, where $t_\mathrm{parallel} =  \SI{15}{\second}$ is the approximate time spent in the parallel phase. The line between the first and second waypoints is parallel to the orientation of the ego vessel when the evasive maneuver is triggered.

\vspace{-3mm}
\subsection{Baseline monitoring approaches}
\label{sec:baselines}
We compare against two alternative monitoring formulations. The first is an instantaneous monitor based on the current states of the vessels. It assesses the \ac{tcpa} \cite{Munoz2016-TCPA} and evaluates the $\mathtt{pos\_encounter}$ predicate to differentiate crossing and head-on encounters. Note that monitoring based on TCPA (or similar risk metrics) and relative position is the most common choice in maritime navigation literature \cite{vagale2021survey}. The second approach uses standard \ac{stl} for monitoring, where the predicted trajectory of the other vessel is given by the centers of the data-driven reachable sets. 
\vspace*{-5mm}
\subsection{Real-world and simulation setup}
For real-world experimentation, we use a model tugboat~(S) as the ego vessel and a drillship (L) as the other vessel. 
The reduced state $\delta_t$ of both vessels is obtained through the motion capture system Qualisys, with approximately $\SI{4}{\meter}$ by $\SI{7}{\meter}$ coverage, limiting the maneuvering and encounter possibilities for the two vessels. 
To introduce environmental disturbance, waves are generated using the JONSWAP (Joint North Sea Wave Project) spectrum \cite{Hasselmann1973JONSWAP}, which models the distribution of wave energy across directions and frequencies. 
We use a peak enhancement factor of 3.5, a peak period of $\SI{1.03}{\second}$, and vary the wave height between $\SI{0.025}{\meter}$ and $\SI{0.035}{\meter}$.
For the simulation experiments, we replace the vessels with a Python-based simulation that uses the dynamics specified in \eqref{eq:boat}, where wave and current effects are neglected. 

\vspace{-1mm}
\section{Experimental Results}
\label{sec:results} 
We evaluate the efficiency, effectiveness, and scalability of \pacstl{} for maritime navigation on a physical testbed in two experimental configurations and compare its performance against \ac{stl} and \ac{tcpa} (Sec.~\ref{sec:baselines}) on \ac{colregs} monitoring tasks. Additionally, we conduct simulation-based ablation studies to assess \pacstl{}'s robustness to vessel types and evaluation parameters, and check for distribution alignment between the reachable set calculation and real-world system.

\subsection{Baseline Comparison}

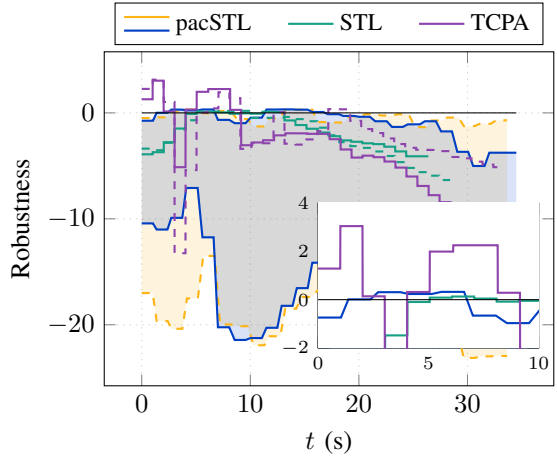
\begin{figure}
    \centering
    \definecolor{pacwaves}{rgb}{0.992, 0.71, 0.082}
\definecolor{pacnowaves}{RGB}{0, 63, 195}
\definecolor{stlcolor}{HTML}{16A085}  
\definecolor{tcpacolor}{HTML}{8E44AD}  

\begin{tikzpicture}
\begin{axis}[
  name        = mainplot,
  width       = 0.85\linewidth,
  height      = 6cm,
  xlabel      = {$t$ (s)},
  ylabel      = {Robustness},
  legend columns = 3,
  legend style = {
    at={(0.5,1.02)},
    anchor=south,
    font=\small,
    /tikz/every even column/.append style={column sep=0.4cm},
  },
  grid        = major,
  grid style  = {dotted, gray!40},
]
  \addplot [name path=mmgraph_pacw_hi, pacwaves, dashed, thick, forget plot]
    table [x=time, y=max_high, col sep=comma] {Data/mm_pacSTL_waves.csv};
  \addplot [name path=mmgraph_pacw_lo, pacwaves, thick, dashed, forget plot]
    table [x=time, y=min_low, col sep=comma] {Data/mm_pacSTL_waves.csv};
  \addplot [pacwaves, opacity=0.15, forget plot]
    fill between [of=mmgraph_pacw_hi and mmgraph_pacw_lo];
  \addplot [name path=mmgraph_pacnw_hi, pacnowaves, thick, forget plot]
    table [x=time, y=max_high, col sep=comma] {Data/mm_pacSTL_no_waves.csv};
  \addplot [name path=mmgraph_pacnw_lo, pacnowaves, thick, forget plot]
    table [x=time, y=min_low, col sep=comma] {Data/mm_pacSTL_no_waves.csv};
  \addplot [pacnowaves, opacity=0.15, forget plot]
    fill between [of=mmgraph_pacnw_hi and mmgraph_pacnw_lo];
  \addplot [stlcolor, thick, dashed, forget plot]
    table [x=time, y=max_ho, col sep=comma] {Data/mm_STL_waves.csv};
  \addplot [stlcolor, thick, forget plot]
    table [x=time, y=max_ho, col sep=comma] {Data/mm_STL_no_waves.csv};
  \addplot [tcpacolor, thick, dashed, forget plot]
    table [x=time, y=max_ho, col sep=comma] {Data/mm_TCPA_waves.csv};
  \addplot [tcpacolor, thick, forget plot]
    table [x=time, y=max_ho, col sep=comma] {Data/mm_TCPA_no_waves.csv};

    \addplot [black, thin, forget plot] 
        table [x=time, y expr=0, col sep=comma] {Data/mm_pacSTL_no_waves.csv};

    \addlegendimage{legend image code/.code={
      \draw[pacnowaves, thick] (0cm,-0.09cm) -- (0.6cm,-0.09cm);
      \draw[pacwaves, thick]   (0cm, 0.09cm) -- (0.6cm, 0.09cm);
    }}
    \addlegendentry{pacSTL}
    \addlegendimage{stlcolor, thick}
    \addlegendentry{STL}
    \addlegendimage{tcpacolor, thick}
    \addlegendentry{TCPA}
\end{axis}

        \begin{axis}[
            at={(mainplot.north east)},
            xshift=-0.2cm,
            yshift=-2cm,
            anchor=north east,
            width=4.5cm,
            height=3.5cm,           
            xmin=0, xmax=10,
            ymin=-2, ymax=4,
            xtick={0,5,10},
            ytick={-2,0,2,4},
            grid=major,
            grid style={dotted, gray!40},
            tick label style={font=\scriptsize},
            title style={font=\scriptsize},
            axis background/.style={fill=white, fill opacity=1},
            axis line style={draw=black, line width=0.4pt},
            clip=true,
        ]
    \fill [white] (rel axis cs:0,0) rectangle (rel axis cs:1,1);
    

    \addplot [name path=inset_pacnw_hi, pacnowaves, thick]
        table [x=time, y=max_high, col sep=comma] {Data/mm_pacSTL_no_waves.csv};
    \addplot [name path=inset_pacnw_lo, pacnowaves, thick, forget plot]
        table [x=time, y=min_low, col sep=comma] {Data/mm_pacSTL_no_waves.csv};
    \addplot [pacnowaves, opacity=0.15, forget plot]
        fill between [of=inset_pacnw_hi and inset_pacnw_lo];


    \addplot [stlcolor, thick]
        table [x=time, y=max_ho, col sep=comma] {Data/mm_STL_no_waves.csv};


    \addplot [tcpacolor, thick]
        table [x=time, y=max_ho, col sep=comma] {Data/mm_TCPA_no_waves.csv};

    \addplot [black, thin, forget plot]
        table [x=time, y expr=0, col sep=comma] {Data/mm_pacSTL_no_waves.csv};

\end{axis}

\end{tikzpicture}
    \caption{Worst-case robustness over time for head-on encounters, comparing pacSTL (blue/yellow), \ac{stl} (teal), and \ac{tcpa} (purple). Solid and dashed lines denote still-water and wave conditions, respectively. The inset highlights the first $\SI{10}{\second}$ for the worst-case, still-water head-on encounter. pacSTL provides earlier detection than \ac{stl}, while \ac{tcpa} triggers inconsistently. }
    \label{fig:methodcomparison}
    \vspace{-3mm}
\end{figure}

Three monitoring techniques (pacSTL, \ac{stl}, and \ac{tcpa}) are compared across two initial configurations, each with ten trials in wave and still-water conditions. \footnote{Video of experiments: https://youtu.be/lQIh8rX26ug} Although the two configurations are designed to generate head-on and crossing encounters, respectively, uncertainty in the vessels' initial orientations leads to varying observed encounter types. The observed shares are presented in the last column of Table~\ref{tab:simrealeval}. Crossing encounters occur more frequently due to the relatively narrow \ac{colregs} sector defining head-on encounters.

Table~\ref{tab:simrealeval} reports the time to evasion $t_e$, defined as the time at which $\overline{h}$ in specification \eqref{eq:persistenthead} or \eqref{eq:persistentcross} first becomes positive, as well as collision and detection rates, where the latter measures the fraction of encounters the monitoring method identifies. Compared with \ac{stl}, \pacstl{} consistently initiates evasive maneuvers earlier (i.e., smaller $t_e$) resulting in lower collision rates. This improvement is especially pronounced in the presence of environmental disturbance (i.e., waves).
\ac{tcpa} achieved lower collision rates and higher detection rates than \ac{stl}, but its behavior is more sensitive to the ego vessel's initial orientation. Therefore, \ac{tcpa} displayed inconsistent behavior with both premature encounter triggers and delayed reactions, resulting in collision and missed detection. 

This is illustrated in Fig.~\ref{fig:methodcomparison}, which compares the worst-case robustness values of each method. pacSTL provides an earlier and more consistent indication of encounter, whereas \ac{stl} detects the encounter several seconds later, and \ac{tcpa} exhibits substantial variability. 
Although waves introduce additional uncertainty, the qualitative results are consistent across still-water and wave conditions.

\begin{table*}[tb]
\caption{Detection Times and Performance Results across Monitoring Strategies}
\begin{center}
\footnotesize
\begin{tabular}{lcccccccc}
\toprule
& \multicolumn{2}{c}{\textbf{Average $t_{e}$}} & \multicolumn{2}{c}{\textbf{Collision Rate $\downarrow$}} & \multicolumn{2}{c}{\textbf{Detection Rate $\uparrow$}} & \multicolumn{2}{c}{\textbf{$\#$ Observed Encounters}}\\
\midrule
\textit{Head-on } & no waves & waves & no waves & waves & no waves & waves & no waves & waves\\
\midrule
\textbf{TCPA} & {$\SI{4.00}{\second}$} & {$\SI{2.95}{\second}$} & 0.00 & 0.10 & 1.00 & 0.90 & 12 & 10 \\
\textbf{STL} & {$\SI{7.80}{\second}$} & {$\SI{8.64}{\second}$} & 0.00 & 0.38 & 0.50 & 0.75 & 6 & 8\\
\textbf{pacSTL} & {$\SI{4.70}{\second}$} & {$\SI{5.54}{\second}$} & 0.00 & 0.00 & 1.00 & 1.00 & 8 & 8\\
\midrule
\textit{Crossing}  \\
\midrule
\textbf{TCPA} & {$\SI{4.95}{\second}$} & {$\SI{4.89}{\second}$} & 0.00 & 0.00 & 1.00 & 1.00 & 8 & 10 \\
\textbf{STL} & {$\SI{6.18}{\second}$} & {$\SI{5.50}{\second}$} & 0.07 & 0.08 & 1.00 & 1.00 & 14 & 12\\
\textbf{pacSTL} & {$\SI{4.04}{\second}$} & {$\SI{5.61}{\second}$} & 0.00 & 0.00 & 1.00 & 1.00 & 12 & 12\\
\bottomrule
\end{tabular}
\label{tab:simrealeval}
\end{center}
\end{table*}

\vspace{-4mm}
\subsection{Real-World Evaluation of Maritime pacSTL}
\begin{figure}
    \centering
    \usetikzlibrary{intersections, pgfplots.fillbetween}

\definecolor{pacwaves}{rgb}{0.992, 0.71, 0.082}
\definecolor{pacnowaves}{RGB}{0, 63, 195}

\begingroup
\begin{tikzpicture}
\begin{axis}[
  width  = 0.85\linewidth,
  height = 6cm,
  xlabel = {$t$ (s)},
  ylabel = {Head-On Robustness},
  legend columns = 2,
  legend style = {
    at={(0.5,1.02)},
    anchor=south,
    font=\small,
    /tikz/every even column/.append style={column sep=0.4cm},
  },
  grid = major,
  grid style = {dotted, gray!40},
]

  \addplot [name path=ho_w_hi_plus,  pacwaves, thick]
    table [x=time, y=high_mean_plus_std,  col sep=comma] {Data/headon_waves.csv};
  \addplot [name path=ho_w_hi_minus, pacwaves, thick, forget plot]
    table [x=time, y=low_mean_minus_std, col sep=comma] {Data/headon_waves.csv};
  \addplot [pacwaves, opacity=0.15, forget plot]
    fill between [of=ho_w_hi_plus and ho_w_hi_minus];
\addlegendentry{Waves}

  \addplot [name path=ho_w_hi, draw=none, forget plot]
    table [x=time, y=high_mean, col sep=comma] {Data/headon_waves.csv};
  \addplot [name path=ho_w_lo, draw=none, forget plot]
    table [x=time, y=low_mean,  col sep=comma] {Data/headon_waves.csv};
  \addplot [pacwaves, opacity=0.15, forget plot]
    fill between [of=ho_w_hi and ho_w_lo];

  \addplot [name path=ho_nw_hi_plus,  pacnowaves, thick]
    table [x=time, y=high_mean_plus_std,  col sep=comma] {Data/headon_no_waves.csv};
  \addplot [name path=ho_nw_hi_minus, pacnowaves, thick, forget plot]
    table [x=time, y=low_mean_minus_std, col sep=comma] {Data/headon_no_waves.csv};
  \addplot [pacnowaves, opacity=0.15, forget plot]
    fill between [of=ho_nw_hi_plus and ho_nw_hi_minus];
\addlegendentry{No Waves}

  \addplot [name path=ho_nw_hi, draw=none, forget plot]
    table [x=time, y=high_mean, col sep=comma] {Data/headon_no_waves.csv};
  \addplot [name path=ho_nw_lo, draw=none, forget plot]
    table [x=time, y=low_mean,  col sep=comma] {Data/headon_no_waves.csv};
  \addplot [pacnowaves, opacity=0.15, forget plot]
    fill between [of=ho_nw_hi and ho_nw_lo];

    \addplot [black, thin, forget plot] 
        table [x=time, y expr=0, col sep=comma] {Data/headon_waves.csv};
\end{axis}
\end{tikzpicture}
\endgroup
    \vspace{0.2cm}
    \usetikzlibrary{intersections, pgfplots.fillbetween}
\definecolor{pacwaves}{rgb}{0.992, 0.71, 0.082}
\definecolor{pacnowaves}{RGB}{0, 63, 195}
\begingroup
\begin{tikzpicture}
\begin{axis}[
  width  = 0.85\linewidth,
  height = 6cm,
  xlabel = {$t$ (s)},
  ylabel = {Crossing Robustness},
  grid = major,
  grid style = {dotted, gray!40}]

  \addplot [name path=cr_w_hi_plus, pacwaves, thick]
    table [x=time, y=high_mean_plus_std,  col sep=comma] {Data/crossing_waves.csv};
  \addplot [name path=cr_w_hi_minus, pacwaves, thick, forget plot]
    table [x=time, y=low_mean_minus_std, col sep=comma] {Data/crossing_waves.csv};
  \addplot [pacwaves, opacity=0.15, forget plot]
    fill between [of=cr_w_hi_plus and cr_w_hi_minus];

  \addplot [name path=cr_w_hi, draw=none, forget plot]
    table [x=time, y=high_mean, col sep=comma] {Data/crossing_waves.csv};
  \addplot [name path=cr_w_lo, draw=none, forget plot]
    table [x=time, y=low_mean,  col sep=comma] {Data/crossing_waves.csv};
  \addplot [pacwaves, opacity=0.15, forget plot]
    fill between [of=cr_w_hi and cr_w_lo];

  \addplot [name path=cr_nw_hi_plus,  pacnowaves, thick]
    table [x=time, y=high_mean_plus_std,  col sep=comma] {Data/crossing_no_waves.csv};
  \addplot [name path=cr_nw_hi_minus, pacnowaves, thick, forget plot]
    table [x=time, y=low_mean_minus_std, col sep=comma] {Data/crossing_no_waves.csv};
  \addplot [pacnowaves, opacity=0.15, forget plot]
    fill between [of=cr_nw_hi_plus and cr_nw_hi_minus];

  \addplot [name path=cr_nw_hi, draw=none, forget plot]
    table [x=time, y=high_mean, col sep=comma] {Data/crossing_no_waves.csv};
  \addplot [name path=cr_nw_lo, draw=none, forget plot]
    table [x=time, y=low_mean,  col sep=comma] {Data/crossing_no_waves.csv};
  \addplot [pacnowaves, opacity=0.15, forget plot]
    fill between [of=cr_nw_hi and cr_nw_lo];

    \addplot [black, thin, forget plot] 
        table [x=time, y expr=0, col sep=comma] {Data/crossing_waves.csv};
\end{axis}
\end{tikzpicture}
\endgroup
    \caption{Smoothed mean of pacSTL robustness intervals over time (dark, inner band), with $\pm 1$ standard deviation (light, outer band) in still-water conditions [blue] and waves [yellow]. Top: 8 head-on encounters. Bottom: 12 crossing encounters.}
    \label{fig:pacstlrobustness}
    \vspace{-4mm}
\end{figure}
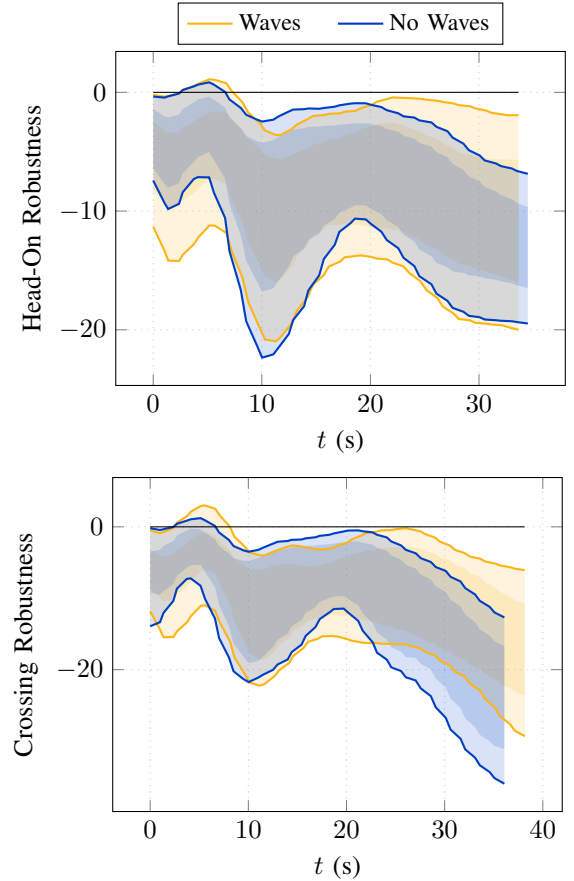

The evolution of the pacSTL robustness intervals during head-on and crossing encounters is shown in Fig.~\ref{fig:pacstlrobustness}. 
Specifically, an encounter is detected when the upper robustness bound, $\overline{h}$, becomes positive, typically around $\SI{4}{}$ or $\SI{5}{\second}$. Once the maneuver is initiated, the robustness interval decreases as the vessel complies with the designated \ac{colregs}. The subsequent increase in robustness after the initial decline is due to the vessel turning to maneuver parallel to its original path. 
Because the reachable tubes account for disturbances, including waves, the larger robustness intervals reflect the increased uncertainty, leading to more robust navigation and decreased risk of collision.  

\subsection{Analysis of Real-World Data-Driven Reachable Sets}
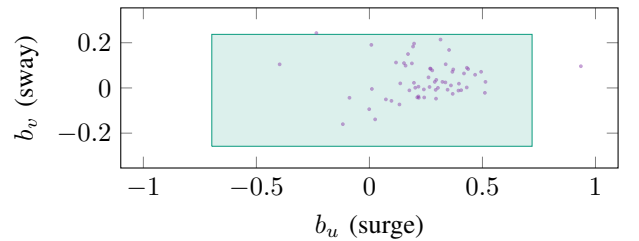
\begin{figure}[b]
    \centering

\definecolor{bdist}{HTML}{16A085}  
\definecolor{outliers}{HTML}{8E44AD}  

\begin{tikzpicture}
\begin{axis}[
    xlabel={$b_u$ (surge)},
    ylabel={$b_v$ (sway)},
    width=.45\textwidth,
    height=3.7cm,
    axis equal,
    xmin=-1.1, 
    xmax=1.1,
    ymin=-0.3, 
    ymax=0.3,
]

\addplot[
    only marks,
    mark=*,
    draw=outliers,
    fill=outliers,
    mark size=0.5pt,
    opacity=0.5,
] table[x=bu, y=bv, col sep=comma] {Data/scatter_waves.csv};

\addplot[
    draw=bdist,
    fill=bdist,
    fill opacity=0.15,
] coordinates {
    (-0.6972786409577466, -0.2580929649597558)
    ( 0.7195866094285801, -0.2580929649597558)
    ( 0.7195866094285801,  0.2369101976929382)
    (-0.6972786409577466,  0.2369101976929382)
    (-0.6972786409577466, -0.2580929649597558)
};

\end{axis}
\end{tikzpicture}
    \caption{Measured disturbance $b_\mathrm{test}$ in surge and sway from $60$ wave trials (purple points), plotted against the $b$-component of the support $S_{\mathcal{D}}$ used for offline, reachable-set computation (teal box).} 
    \label{fig:bdistribution}
\end{figure}

To assess whether the offline-computed reachable sets, which incorporate experimentally-derived disturbance $b$, remain valid during real-world operation, we perform an a posteriori analysis of the disturbances encountered during experimentation. During the experiments, the distribution of control inputs $\tau$ is sampled identically to the offline reachable set computation. The realized control inputs are logged and used to compute the disturbance term $b_\mathrm{test}$. 
Specifically, we analyze the $60$ trials conducted under wave conditions and compare the measured disturbances $b_\mathrm{test}$ against the support $S_{\mathcal{D}}$ used for offline reachability analysis. As shown in Fig. ~\ref{fig:bdistribution}, only two of the $60$ trials fall outside the assumed support.
Importantly, the reachable sets are constructed using the scenario approach, which emphasizes worst-case system behavior. Consequently, occasional violations of the support does not necessarily imply reachable set violations.

\subsection{Ablation Studies}
\textbf{Specification Complexity:} To demonstrate the expressiveness of pacSTL, we analyze the effect of increasing the specification complexity:
\begin{equation}\label{eq:persistentencounter-morecomplex}
    \Phi \eqqcolon \lnot \mathtt{encounter} \land \mathrm{G}_{[t_\mathrm{start}, t_\mathrm{end}]} (\mathtt{encounter}).
\end{equation}
The leading negated term ensures that the specification is satisfied only shortly before the onset of a persistent encounter, providing a more refined trigger for initiating an evasive maneuver. This transition-based specification is more challenging, as detecting the onset is more sensitive to low observation frequencies. Therefore, the increased complexity resulted in only sporadic detection when using \ac{stl}. Nevertheless, \pacstl{} maintained an average time to evasion of $t_e=\SI{6.7}{\second}$, compared to $t_e=\SI{6.8}{\second}$ for the original specification, with a $100\%$ detection rate.

\textbf{Time Horizon:} The time horizon parameter $t_h$ governs the behavior of atomic proposition $h_\mathtt{collision\_risk}$. Higher values of $t_h$ increase the conservativeness of \pacstl{}, leading to earlier times of detection $t_e$. We tested $t_h=10$ (default value) and $t_h=20$, leading to average respective times of evasion $t_e =\SI{6.8}{\second} $ and $t_e =\SI{4.0}{\second} $.
There is no computational overhead to changing the parameter $t_h$ with \pacstl{} and can even be changed throughout a mission.

\textbf{Vessel Type:} In simulation, we evaluated three different combinations of vessel types (S-L, which is the default on hardware, S-S, and L-S). Incorporating additional vessel types only requires offline computation of their reachable tubes, which can then be reused across scenarios, e.g., S-S and L-S use the same reachable tubes since the other vessel is S in both cases. As expected, when the large vessel is designated as the ego vessel, the average time to evasion increases from $t_e =\SI{6.8}{\second} $ to $t_e =\SI{8.5}{\second}$, due to smaller reachable sets associated with the other vessel, S.

\addtolength{\textheight}{-8mm} 

\section{Discussion}
\label{sec:discussion}
Our experiments demonstrate that uncertainty-aware monitoring improves decision making. Compared with \ac{stl} and \ac{tcpa}, pacSTL consistently detects encounters earlier and reduces collision rates.
The ablation studies further show that these improvements persist with increased specification complexity and varying parameterization.
The proposed framework obtains this flexibility by shifting the computational burden largely offline through data-driven reachable sets. 
During deployment, monitoring only requires evaluating the pacSTL optimization problems presented in Sec. \ref{sec:maritime_nav}, which takes approximately 0.15 - 0.6 seconds using a naive implementation based on standard Python packages (achieving real-time operation for maritime navigation where 1-2Hz is expected). 
To reduce runtime for more dynamic robotic applications, the optimization problems for the atomic propositions could be parallelized and reused. Additionally, deriving closed-form solutions for linear atomic propositions would reduce computational overhead, motivating the use of specifications primarily composed of linear atomic propositions.

In our experiments, uncertainty in the state estimation of the ego vessel can be neglected due to the high accuracy of the motion capture system. Additionally, in full-scale applications, state estimates are typically relative to the ego vessel, which implicitly incorporates ego uncertainty into the reachable sets of the other vessel. 
Nevertheless, developing pacSTL formulations that account for reachable sets of multiple agents would broaden the applicability of pacSTL. 
Extending the framework to multiple agents requires optimization problems for atomic robustness functions that efficiently determine valid lower and upper bounds and 
deriving \ac{pac} guarantees for specifications involving multiple \ac{pac}-bounded reachable sets.

\section{Conclusion}
\label{sec:conclude}
We propose an uncertainty-aware, real-time monitoring framework based on pacSTL. By combining experimentally derived distribution models with high-fidelity simulations, the framework enables data-driven reachability analysis for real-world robotic systems. 
Experimental results in maritime navigation demonstrate that the framework provides more robust monitoring than simpler metrics in the presence of environmental disturbances, consistently detecting encounters and avoiding collisions in real-world experiments. This work highlights the potential of pacSTL for deployment across a wide range of robotic systems.








\bibliography{ref}
\bibliographystyle{IEEEtran}

\end{document}